%% file: root.tex
\documentclass[conference]{IEEEtran}
\IEEEoverridecommandlockouts    

\usepackage{multirow}
\usepackage{lipsum}
\usepackage{amsmath}
\usepackage{amssymb}
\usepackage[ruled,vlined]{algorithm2e}
\usepackage{graphicx}
\usepackage[utf8]{inputenc}
\usepackage[T1]{fontenc}
\graphicspath{{./Figures/}}
\usepackage{acro}          
\input{acronyms.tex}       

\title{\LARGE \bf Refining Roadside 4D Radar Vehicle Classification Using Point-Cloud Dispersion}

\author{Mohammed Alsakabi$^{1,*}$, Andreia Figueiredo$^{2,3,*}$, António Almeida$^{2,3}$, John Dolan$^{1}$,\\
Miguel Luís$^{4}$, Pedro Rito$^{2,3}$, Ozan Tonguz$^{1}$, Susana Sargento$^{2,3}$%
\thanks{$^{*}$These authors contributed equally to this work.}%
\thanks{$^{1}$Carnegie Mellon University, Pittsburgh, USA}%
\thanks{$^{2}$Departamento de Electrónica, Telecomunicações e Informática, Universidade de Aveiro, Portugal}%
\thanks{$^{3}$Instituto de Telecomunicações, Universidade de Aveiro, Campus Universitário de Santiago, 3810-193, Aveiro, Portugal}%
\thanks{$^{4}$Instituto de Telecomunicações, Instituto Superior Técnico, Universidade de Lisboa, 1049-001 Lisboa, Portugal}%
}

\begin{document}
	\maketitle
	\thispagestyle{empty}
	\pagestyle{empty}

	\input{1_abstract}
    \input{2_intro}
    \input{3_related_work}

\input{4_sensing_platform}

\input{5_dataset_overview}
    \input{6_proposed_improvement_to_classification}

\input{7_discussion}
    \input{8_conclusion}

	
	\bibliographystyle{IEEEtran}
	\bibliography{root} 
	
\end{document}

%% file: acronyms.tex
\DeclareAcronym{gnss}{
    short           = GNSS,
    long            = global navigation satellite system,
}

\DeclareAcronym{fmcw}{
    short           = FMCW,
    long            = Frequency Modulated Continuous Waves,
}

\DeclareAcronym{mimo}{
    short           = MIMO,
    long            = Multiple-Input and Multiple-Output,
}

\DeclareAcronym{fft}{
    short           = FFT,
    long            = Fast Fourier Transform,
}
\DeclareAcronym{its-s}{
    short           = ITS-S,
    long            = Intelligent Transportation System Station,
}
\DeclareAcronym{its}{
    short           = ITS,
    long            = Intelligent Transportation Systems,
}

\DeclareAcronym{radar}{
    short           = radar continental,
    long            = radar continental ARS548RDI,
}

%% file: 1_abstract.tex
\begin{abstract}
Roadside sensing infrastructure is essential for  Intelligent Transportation Systems (ITS), but cameras and legacy sensors are limited in range, weather robustness, and depth accuracy. 4D radar complements these modalities by resolving range, azimuth, elevation, and radial velocity under adverse conditions, though its sparse, noisy point clouds make reliable classification difficult. We present a roadside multi-sensor platform combining a 4D radar and stereo depth camera under synchronized acquisition, used to collect a real-world dataset. Our analysis reveals a substantial class-distribution discrepancy between modalities, most notably a strong radar confusion between passenger cars and motorcycles (49.01\% and\ 48.60\% of the dataset). To address this, we propose a Neyman-Pearson-based binary classifier on the temporally-smoothed 2D positional standard deviation of radar point cloud clusters, reducing classification error from 52.1\% to 42.7\%. This lightweight and edge-deployable approach improves radar-only classification reliability and lays the foundation for future camera-radar fusion.
\end{abstract}

%% file: 2_intro.tex
\section{Introduction}
\ac{its} rely on accurate, real-time perception of traffic to support applications such as signal control, incident detection, and safety analytics. Roadside sensing infrastructure, including cameras, inductive loops, and increasingly radar, provides the situational awareness needed to monitor vehicles, cyclists, and pedestrians at intersections and along road segments \cite{cress2023intelligent, wang2021roadside, guerrero2018sensor}. This need is especially acute at high-traffic locations, where interactions between different road users pose the greatest safety risk, and where reliable perception directly informs signal control and safety analytics.

\par Cameras remain the dominant modality for traffic monitoring due to their low cost and rich semantic output, but they are limited by sensitivity to lighting and weather conditions, occlusion, and constrained depth accuracy, and typically cover only a near-field region around the sensor \cite{guerrero2018sensor, loce2013computer}. Inductive loops and other legacy sensors offer robustness but provide coarse, point-level information with little to no classification or trajectory detail \cite{klein2006traffic}. LiDAR offers precise 3D geometry and longer range, but remains costly, degrades in adverse weather, and lacks direct velocity measurement \cite{kim2023empirical, li2020lidar, roriz2021automotive}. These limitations motivate sensing modalities that can operate reliably over longer ranges and under a wider variety of environmental conditions.

\par 4D radar has emerged as a promising complement to vision-based sensing \cite{fan20244d}. Unlike legacy automotive 3D radar, 4D radar resolves range, azimuth, elevation, and radial velocity, enabling richer point cloud and object-level representations while remaining largely unaffected by lighting and adverse weather. Radar also offers substantially longer sensing range than typical roadside cameras, making it attractive for wide-area traffic monitoring. However, radar-based classification remains challenging: sparse, noisy point clouds and limited geometric expressiveness make it difficult to reliably distinguish between object classes, particularly at longer ranges \cite{zhou2022towards}.

\par In this work, we present a roadside multi-sensor perception platform combining a 4D radar and a stereo depth camera under synchronized acquisition, and use it to collect a dataset under real urban traffic conditions. We analyze the detection and classification characteristics of both modalities, revealing a significant class-distribution discrepancy between radar and camera outputs, most notably a strong radar confusion between passenger cars and motorcycles. To address this, we propose a Neyman-Pearson-based binary classifier operating on the temporally-smoothed positional standard deviation of radar point cloud clusters, and show that temporal averaging substantially improves class separability for our proposed system, reducing classification error by 9.4 percentage points compared to the baseline setup. These results demonstrate a real-world lightweight, edge-deployable improvement to radar-based object classification and lay groundwork for future camera-radar fusion.

\par The paper is organized as follows. Section II reviews related work on 4D radar datasets and radar-based object detection and classification. Section III describes the sensing platform and its radar and camera subsystems. Section IV presents the collected dataset and analyzes the classification discrepancy between modalities. Section V introduces the proposed classifier, and Section VI discusses the results and limitations. Finally, Section VII concludes the paper. 

%% file: 3_related_work.tex
\section{Related Work}
This section presents related work on existing datasets and approaches for detection and classification. 

\subsection{4D Radar-Camera Datasets}
Several datasets support radar-based and radar-camera fusion perception across vehicle-mounted, infrastructure-based, and multimodal settings. CARRADA and RadarScenes established early benchmarks for vehicle-mounted radar, offering dense range-angle-Doppler maps and manually annotated point clouds, respectively \cite{ouaknine2021carrada, schumann2021radarscenes}. RADDet similarly targets vehicle-mounted sensing, introducing Range-Azimuth-Doppler tensors with an auto-annotation pipeline for road-user detection \cite{zhang2021raddet}. For richer multimodal fusion, RaDelft, View of Delft, and K-Radar provide synchronized 4D radar, lidar, camera, and odometry data \cite{roldan2024deep, palffy2022multi, paek2022k}, while nuScenes and RADIATE offer large-scale, multi-sensor coverage across diverse driving scenes and adverse weather \cite{caesar2020nuscenes, sheeny2021radiate}. All of these 4D radar datasets are collected from vehicle-mounted platforms.

Fewer datasets target infrastructure-based sensing. Radar Lab and the City of Austin Radar Traffic Counts dataset provide open traffic monitoring data, though the former offers annotated frames for classification, while the latter reports only aggregate counts and speeds \cite{soom2024open, austin_radar_traffic_counts}. INFRA-3DRC is the closest match, providing synchronized 3D radar and RGB camera frames from a roadside setup with semi-automatic point-wise and bounding-box annotations \cite{agrawal2023infra}, but relies on 3D rather than 4D radar. To our knowledge, no existing dataset combines 4D radar with a roadside infrastructure viewpoint; existing 4D radar work remains vehicle-mounted, while infrastructure-based datasets rely on 3D radar or aggregate traffic statistics. Our dataset addresses this gap by pairing 4D radar with camera images from a fixed roadside setup.

\subsection{4D Radar Object Detection and Classification}
4D radar-only detection methods have progressively addressed the core challenges of point cloud sparsity, noise, and limited geometric expressiveness. Early work adapted pillar-based LiDAR methods to 4D radar. RPFA-Net replaces PointNet with a self-attention mechanism to capture global inter-pillar context for improved heading angle regression \cite{xu2021rpfanet}. SMURF combines pillarization with kernel density estimation (KDE) to suppress noise and capture spatial density distributions \cite{liu2024smurf}. MUFASA enriches features through Lalonde-descriptor-based geometric analysis and a distributed multi-view attention module integrating dataset-wide shared features across BEV and cylindrical projections \cite{peng2024mufasa}. MAFF-Net combines sparsity pillar attention, velocity-clustering-guided cross-attention, and cylindrical denoising to jointly address sparsity and noise  \cite{bi2025maffnet}. Radar M$^3$-Net expands the receptive field via multi-scale voxel encoding and multi-layer attention, with temporal BEV fusion and a sparse dual-head for large object detection \cite{yang2025radarm3net}. Wang~\textit{et al.}~\cite{wang2026dual} address sparsity via a dual-view diffusion framework that reconstructs BEV and front-view depth maps, fusing the resulting pseudo points with raw radar features using deformable cross-attention. RadarNeXt, a lightweight detector, combines a re-parameterizable backbone with a multi-path foreground enhancement network, achieving over 50 mAP on View-of-Delft at more than 67 FPS \cite{jia2025radarnext}. Unlike these approaches, which rely on deep architectures trained on dense, vehicle-mounted point clouds, our method targets the car–motorcycle confusion specific to sparse roadside deployments using a single lightweight, interpretable feature and a Neyman-Pearson threshold, trading general multi-class capacity for low-cost edge deployability.

%% file: 4_sensing_platform.tex
\section{Sensing Platform}

\begin{figure}[h]
    \centering
    \includegraphics[width=0.75\linewidth]{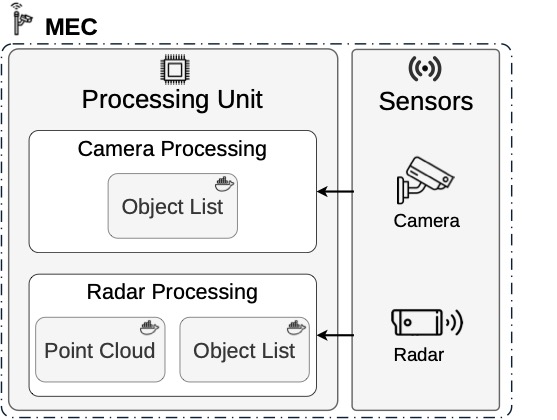}
    \caption{Architecture of the sensing platform.}
    \label{fig:ArchitectureSensingPlataform}
\end{figure}

Multi-sensor perception, aligned with fusion mechanisms, enables \ac{its} to combine complementary sensing modalities and improve the understanding of the surrounding environment. To obtain data under real traffic conditions and enable the evaluation of such approaches, a roadside multi-sensor perception platform was designed to acquire a dataset with a radar and a camera operating under synchronized acquisition (Figure~\ref{fig:ArchitectureSensingPlataform}). The system is designed around two parallel data-acquisition pipelines, one for radar and one for camera, each with independent processing and storage.

\subsection{Radar Subsystem}

The sensing platform employs the ARS548 RDI Continental radar, a long-range 4D radar designed for automotive and infrastructure applications. The sensor operates using \ac{fmcw} chirps transmitted through a \ac{mimo} antenna array. Within the radar’s embedded signal-processing pipeline, the received echoes are processed through proprietary multi-dimensional \acp{fft} and detection stages to estimate range, radial velocity, azimuth, and elevation for each reflection point. These object observations are exported as a 4D point cloud. In parallel, the radar’s internal tracking pipeline performs clustering and target association, aggregating reflection points into discrete dynamic objects. Each reported object includes kinematic state variables such as position, velocity, acceleration, heading, object dimensions, and a classification label with associated confidence.

The radar exports both the 4D point cloud and the tracked object list through a UDP multicast interface. To capture these data streams, a custom software stack developed in-house was deployed on the edge node as a Docker container, corresponding to the Radar Processing module shown in Figure~\ref{fig:ArchitectureSensingPlataform}. The application subscribes to the multicast group and continuously records the incoming radar data in real time. The incoming radar outputs are divided into two independent processing and storage pipelines: a Point Cloud pipeline, which processes and stores reflection-level detections, and an Object List pipeline, which processes and stores tracked objects.

\subsubsection{\textbf{Point Cloud pipeline}}

The Point Cloud pipeline receives the radar point cloud output at 20 Hz. Each exported frame contains 800 point entries, corresponding to the fixed-size point cloud structure provided by the Continental ARS548 RDI radar. Valid entries represent detected reflections, whereas unused entries are zero-filled. For each valid object observation, the sensor reports five attributes: range, azimuth, elevation angle, radial velocity (range rate), and radar cross-section (RCS). 

To facilitate downstream processing, the native polar representation is converted into a local three dimensional Cartesian coordinate system $(x, y, z)$. This transformation relies on the measured range and angles, computed using the following trigonometric relations:
\begin{align}
x &= -\,r \sin(\theta)\cos(\phi) \\
y &= r \cos(\theta)\cos(\phi) \\
z &= r \sin(\phi)
\end{align}

where $r$ denotes range, $\theta$ azimuth, and $\phi$ elevation.

Each processed frame is stored as an individual Point Cloud Data (\texttt{.pcd}) file, sequentially named \texttt{point\_sweepX.pcd}. Each file corresponds to a single timestamped radar observation and contains both the raw radar measurements and the derived Cartesian coordinates for all 800 points.
To preserve the acquisition time associated with each frame, the recording software appends a reception timestamp at the moment the data are captured. Since standard PCL custom fields use 32-bit floating-point storage, the native 64-bit epoch timestamp cannot be represented directly without precision loss. To maintain microsecond resolution, the timestamp, $t$, is divided into two fields, $t_{high}$ and $t_{low}$, using a 20-bit split:
\begin{align}
t_{high} &= t \gg 20 \\
t_{low} &= t \,\&\, 0\text{xFFFFF}
\end{align}

where $t_{\text{low}}$ selects the lower 20 bits of the original timestamp, while $t_{\text{high}}$ stores the remaining upper bits. The complete timestamp in milliseconds can later be reconstructed during data loading as
\begin{align}
t =
(t_{high}\ll 20) + t_{low}
\end{align}

The attributes stored in each \texttt{.pcd} file are summarized in Table~\ref{tab:pcd_fields}.

\begin{table}[h]
\centering
\caption{Point cloud attributes stored in each \texttt{.pcd} file.}
\label{tab:pcd_fields}
\begin{tabular}{lll}
\hline
\textbf{Field} & \textbf{Description} & \textbf{Unit} \\ \hline
x & Cartesian X coordinate & m \\
y & Cartesian Y coordinate & m \\
z & Cartesian Z coordinate & m \\
range & Measured target distance & m \\
azimuth & Horizontal angle & rad \\
elevation & Vertical angle & rad \\
rcs & Radar cross section & dBm$^2$ \\
range\_rate & Radial velocity & m/s \\
timestamp\_high & Upper timestamp segment & -- \\
timestamp\_low & Lower timestamp segment & -- \\ \hline
\end{tabular}
\end{table}

\subsubsection{\textbf{Object List pipeline}}

In parallel with the Point Cloud stream, the radar additionally exports a tracked Object List generated by its internal clustering and tracking pipeline. This output represents processed object tracks rather than raw radar reflections, with a maximum of 50 tracked objects reported per update cycle. Each detected object is assigned a persistent identifier, enabling the temporal reconstruction of trajectories across consecutive radar updates.

For every tracked object, the radar reports a set of semantic, geometric, and kinematic attributes. These include the object identifier, classification label, classification confidence, local Cartesian position $(x,y)$, speed, acceleration, heading, and estimated object length. In addition, the radar provides uncertainty estimates in the form of standard deviations for position, speed, acceleration, and heading, allowing downstream algorithms to assess the reliability of each state estimate.

Object positions are initially expressed in the radar-centric local reference frame. To facilitate geographic visualization and multi-sensor fusion, these local coordinates are additionally converted into global latitude and longitude coordinates. The transformation uses the known installation parameters of the radar, namely its geographic anchor position and mounting heading. The heading reference is adjusted such that the radar forward direction is aligned with geographic North, ensuring that all converted object observations are represented in a consistent North--East reference frame.

After acquisition, each object report is processed individually by the processing service and stored in a SQLite database. Each row corresponds to a single tracked-object observation at a specific timestamp and contains both local radar-centric coordinates and global geographic coordinates, together with the associated kinematic states, classification estimates, and uncertainty measures. The complete list of stored attributes is summarized in Table~\ref{tab:radar_object_fields}.

\begin{table}[ht]
\centering
\caption{Attributes stored in the radar SQLite Object List database.}
\label{tab:radar_object_fields}
\scriptsize
\setlength{\tabcolsep}{3pt}
\begin{tabular}{p{2.0cm} p{4.1cm} p{0.8cm}}
\hline
\textbf{Field} & \textbf{Description} & \textbf{Unit} \\ \hline
timestamp & Object observation timestamp (UNIX epoch) & s \\
objectID & Unique radar tracking identifier & -- \\
classification & Object class label (1=person, 2=bicycle, 4=motorcycle, 5=car, 6=bus, 7=truck) & -- \\
confidence & Classification confidence & \% \\
x & Local X position in radar frame & m \\
y & Local Y position in radar frame & m \\
xStd & Standard deviation of X estimate & m \\
yStd & Standard deviation of Y estimate & m \\
latitude & Global latitude of object & deg \\
longitude & Global longitude of object & deg \\
speed & Estimated object speed & m/s \\
speedStd & Standard deviation of speed estimate & m/s \\
acceleration & Estimated object acceleration & m/s$^2$ \\
accelerationStd & Standard deviation of acceleration estimate & m/s$^2$ \\
heading & Estimated object heading & deg \\
headingStd & Standard deviation of heading estimate & deg \\
length & Estimated object length & m \\ \hline
\end{tabular}
\end{table}

\subsection{Camera Subsystem}
The visual sensing component of the platform is a Luxonis OAK 4D, a standalone edge-inference stereo depth camera built around a 48~MP RGB sensor and a pair of 1~MP global-shutter stereo sensors separated by a 7.5~cm baseline. The device integrates a six-core ARM processor, 8~GB of RAM, and a neural processing unit capable of up to 52 TOPS, enabling on-device perception without external acceleration.

The camera internally acquires synchronized RGB imagery from the primary sensor together with a rectified stereo pair from the global-shutter sensors. Within the embedded processing pipeline, the stereo streams are used for disparity estimation and depth reconstruction, while the RGB stream is processed by an onboard neural network through the Luxonis DepthAI framework to perform object detection. The resulting tracked object list is exported by the camera as the primary output used in this work.

To capture these outputs, a custom software stack developed in-house was deployed on the edge node as a Docker container, corresponding to the Camera Processing module shown in Fig.~\ref{fig:ArchitectureSensingPlataform}. 

\subsubsection{\textbf{Object List Pipeline}}

For each object observation, the camera exports a semantic class label, confidence score, and three-dimensional local Cartesian position $(x,y,z)$ derived from the onboard depth estimation pipeline. A persistent identifier is also associated with each object, enabling the temporal reconstruction of trajectories across consecutive frames.

The exported observations are received in real time by the in-house acquisition service running on the edge node. After the reception, each object observation is processed. Object positions, initially expressed in the camera-centric local reference frame, are transformed into global latitude and longitude coordinates using the known installation parameters of the camera, namely its geographic anchor position and mounting heading. The heading reference is adjusted such that the camera forward direction is aligned with geographic North, ensuring that all converted object observations are represented in a consistent North--East reference frame.

The processed object observations are subsequently stored in a SQLite database. Each row corresponds to an object observation at a specific timestamp and contains the local and global coordinates, along with the associated class label and confidence score. The complete list of stored attributes is summarized in Table~\ref{tab:camera_object_fields}.

\begin{table}[ht]
\centering
\caption{Attributes stored in the camera SQLite Object List database.}
\label{tab:camera_object_fields}
\scriptsize
\setlength{\tabcolsep}{3pt}
\begin{tabular}{p{2.0cm} p{4.1cm} p{0.8cm}}
\hline
\textbf{Field} & \textbf{Description} & \textbf{Unit} \\ \hline
timestamp & Object observation timestamp (UNIX epoch) & s \\
objectID & Unique camera tracking identifier & -- \\
classification & Object class label (1=person, 2=bicycle, 4=motorcycle, 5=car, 6=bus, 7=truck) & -- \\
confidence & Classification confidence & \% \\
latitude & Global latitude of object & deg \\
longitude & Global longitude of object & deg \\
x & Local X position in camera frame & m \\
y & Local Y position in camera frame & m \\
z & Local Z position in camera frame & m \\
reference\_latitude & Anchor Position of camera & deg \\
reference\_longitude & Anchor Position of camera & deg \\
reference\_heading & Mounting Heading & deg \\ \hline
\end{tabular}
\end{table}

%% file: 5_dataset_overview.tex
\section{Dataset Overview}
The dataset was collected during a single daytime acquisition session at a public urban road intersection under clear weather and dry conditions. Data collection was conducted during active weekday traffic hours to capture a representative mix of urban road users, including passenger cars, motorcycles, bicycles, pedestrians, and occasional commercial vehicles.

\begin{figure}[h]
    \centering
    \includegraphics[width=0.75\linewidth]{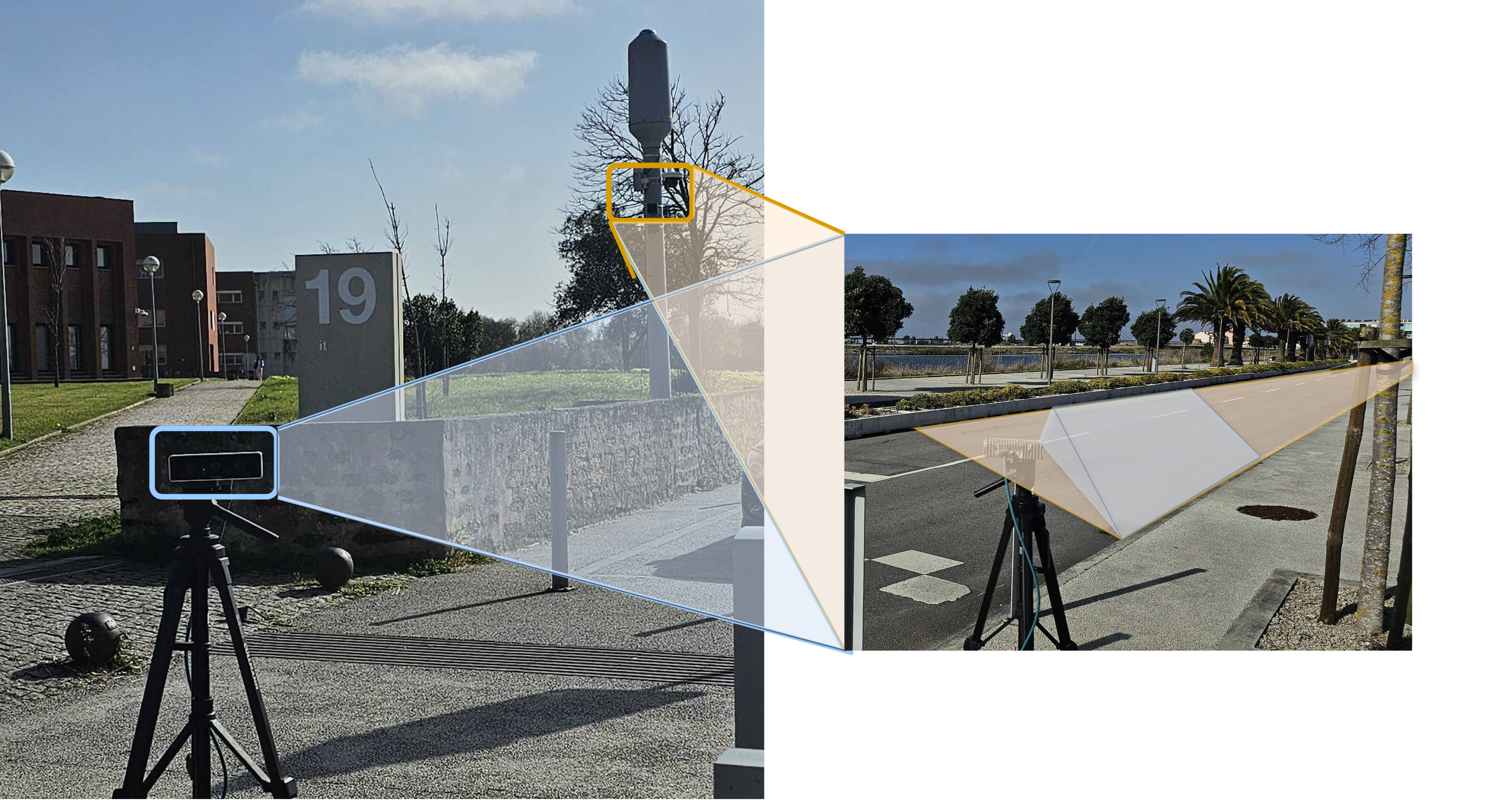}
    \caption{Roadside deployment of the radar and stereo camera.}
    \label{fig:FieldTest}
\end{figure}

Both sensors were co-located at a fixed position overlooking the monitored road. The radar (orange box in Figure~\ref{fig:FieldTest}) was mounted approximately 3~m above ground level, while the camera (blue box in Figure~\ref{fig:FieldTest}) was installed approximately 1~m above ground level. Due to the different sensing ranges of the two modalities, the camera was positioned several meters ahead of the radar to maximize the shared field of view over the monitored traffic lanes, as shown in Figure~\ref{fig:FieldTest}.

To enrich the dataset with long continuous trajectories of a known target, a designated test vehicle was repeatedly driven through the intersection at reduced speed while normal traffic conditions remained unchanged. This procedure enabled the collection of dense radar tracks and point cloud observations under realistic traffic flow.


\subsection{Statistical Overview}

The resulting dataset was subsequently analyzed to characterize the observations from each sensing modality in terms of observations' counts, class distributions, and spatial coverage.

\begin{figure}[h]
    \centering
    \includegraphics[width=1\linewidth]{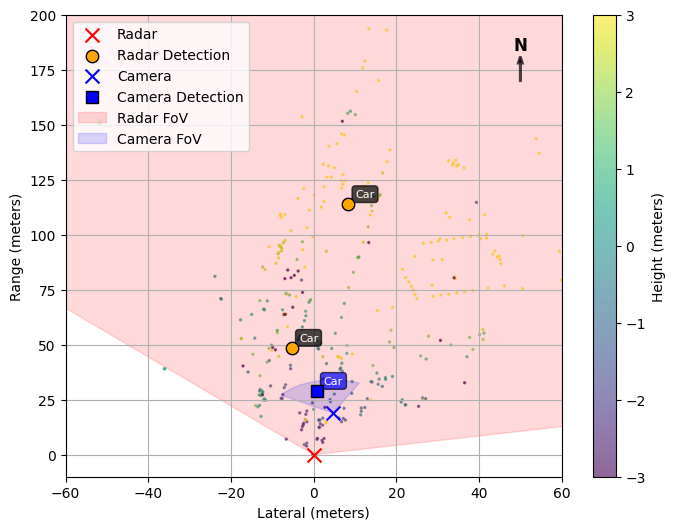}
    \caption{Bird's-eye view of radar and camera object observations}
    \label{fig:BirdEyeView}
\end{figure}

The radar subsystem produced a total of 26,081 validated object observations, substantially exceeding the 3,467 object observations generated by the camera subsystem. This difference is primarily explained by the wider field of view and extended sensing range of the radar, which enables observations over a broader roadway segment than the near-field camera coverage. Figure~\ref{fig:BirdEyeView} illustrates the effective sensing regions of both modalities.

The radar captured extended object trajectories across the monitored intersection, with recorded object observation ranges varying from 11~m to more than 135~m from the sensor origin. In contrast, the camera operated primarily in the near-field region, with object observations between 1.5~m and 13~m from its installation position.

Table~\ref{tab:class_distribution} summarizes the distribution of classified radar object observations. Passenger cars and motorcycles dominate the radar dataset with nearly equal proportions, accounting for 49.01\% and 48.60\% of object observations, respectively. Bicycles represent 2.15\% of observations, while trucks account for only 0.24\%. 

\begin{table}[h]
\centering
\caption{Distribution of Classified Radar object observations}
\label{tab:class_distribution}
\begin{tabular}{lcc}
\hline
\textbf{Object Class} & \textbf{Total object observations} & \textbf{Percentage (\%)} \\ \hline
Passenger Car         & 12,781                    & 49.01                    \\
Motorcycle            & 12,675                    & 48.60                    \\
Bicycle               & 562                       & 2.15                     \\
Truck                 & 63                        & 0.24                     \\ \hline
\textbf{Total}        & \textbf{26,081}           & \textbf{100.00}          \\ \hline
\end{tabular}
\end{table}

The camera object observations, summarized in Table~\ref{tab:camera_class_distribution}, exhibit a different class composition. Passenger cars constitute the majority of object observations (85.69\%), followed by trucks (5.34\%) and pedestrians (5.31\%). Bicycles and motorcycles are comparatively less frequent.

\begin{table}[h]
\centering
\caption{Distribution of Classified Camera object observations}
\label{tab:camera_class_distribution}
\begin{tabular}{lcc}                                          
\hline
\textbf{Object Class} & \textbf{Total object observations} & \textbf{Percentage (\%)} \\ \hline                
Passenger Car         & 2,971                     & 85.69                    \\
Truck                 & 185                       & 5.34                     \\
Person                & 184                       & 5.31                     \\
Bicycle               & 77                        & 2.22                     \\
Motorcycle            & 40                        & 1.15                     \\ \hline
\textbf{Total}        & \textbf{3,467}            & \textbf{100.00}          \\ \hline
\end{tabular}
\end{table}

\subsection{Classification Challenges}

\begin{figure}[h]
    \centering
    \includegraphics[width=0.8\linewidth]{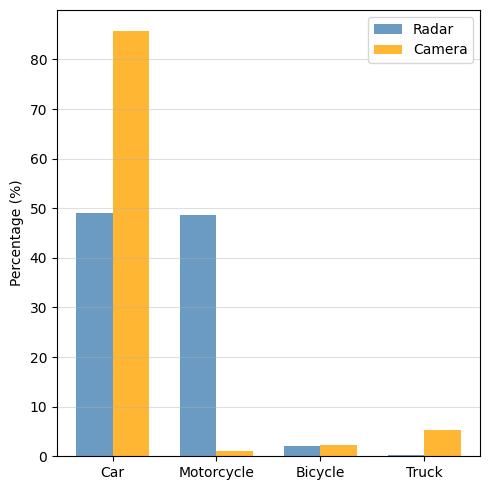}
    \caption{Comparison of percentages between Radar and Camera object observations}
    \label{fig:percent}
\end{figure}

A comparison between Tables~\ref{tab:class_distribution} and~\ref{tab:camera_class_distribution}, together with Fig.~\ref{fig:percent}, reveals a marked discrepancy in the class distributions reported by the two sensing modalities. Although the radar and camera cover different distance ranges, the monitored road segment is straight and contains no entrances or exits between their respective observation regions. Therefore, vehicles observed by the radar at longer distances are expected to subsequently cross the camera’s field of view. The two sensors can consequently be regarded as observing the same underlying traffic stream, although the number of recorded observations differs because of their distinct sensing ranges, update rates, and track durations.

Passenger cars dominate the camera object observations, representing 85.69\% of the total, whereas the radar reports nearly equal proportions of passenger cars and motorcycles, at 49.01\% and 48.60\%, respectively. This unexpectedly high proportion of motorcycle labels is inconsistent with the traffic observed during data collection and with the class composition subsequently reported by the camera. As a result, the radar's class composition should be interpreted with caution, highlighting the need for a more thorough examination of the classification outputs.

\begin{figure}[h]
    \centering
    \includegraphics[width=1\linewidth]{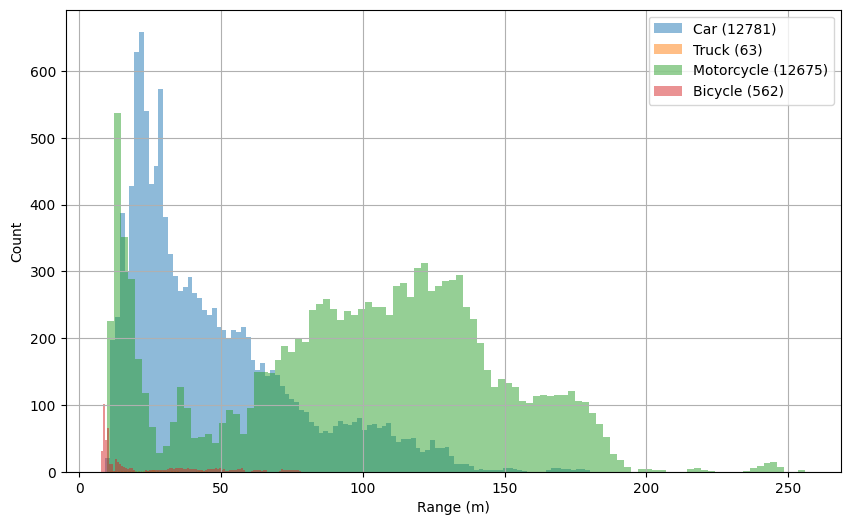}
    \caption{Radar object observations across different ranges}
    \label{fig:radar_ranges}
\end{figure}

Figure~\ref{fig:radar_ranges} further shows that the distribution of radar-assigned classes varies substantially with object observation range. Passenger car observations are concentrated at shorter and intermediate distances, where target shape and spatial extent are more reliably resolved. In contrast, motorcycle labels become increasingly dominant at longer ranges, where reduced angular resolution, lower point density, and weaker geometric cues can affect the onboard classifier.

A possible explanation is that long-range sensing conditions reduce the amount of discriminative information available to the onboard classifier. At extended distances, lower angular resolution, reduced point density, and less reliable geometric extent estimates can make certain vehicle classes harder to distinguish. These limitations motivate the development of an improved classifier to enhance the semantic reliability of radar object observations.

%% file: 6_proposed_improvement_to_classification.tex
\section{Proposed Improvements to Classification}
Building on these observations, the following approach explores how $\sigma$-based features can be used to improve the distinction between passenger cars and motorcycles.

Reference labels for the passenger car and motorcycle classes used throughout this section were taken from the radar's own onboard classification output, as no independent per-object ground truth was available at the ranges where class confusion is most pronounced. The reported error rates should be interpreted as a measure of the separability of the proposed $\sigma$-based feature relative to the radar's own class assignments, rather than as absolute classification accuracy against independent ground truth.

\subsection{Base Classifier}

\begin{figure}[t]
    \centering
    \includegraphics[width=0.75\columnwidth]{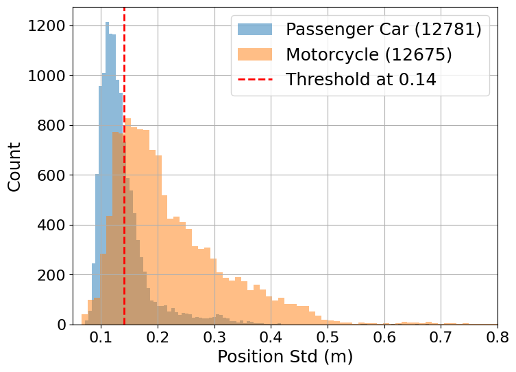}
    \caption{Distribution of 2D positional standard deviation $\sigma$ for 
    passenger cars and motorcycles. The dashed red line indicates the 
    Neyman-Pearson decision threshold $\eta = 0.14$ m.}
    \label{fig:base_classifier}
\end{figure}

The Neyman-Pearson (NP) criterion provides a principled framework for binary hypothesis testing under asymmetric error constraints \cite{tong2016survey}. Given two competing hypotheses $\mathcal{H}_0$ (passenger car) and $\mathcal{H}_1$ (motorcycle), the NP lemma states that the most powerful test at a fixed significance level $\alpha$ is the likelihood ratio test:

\begin{equation}
    \Lambda(\sigma) = \frac{p(\sigma \mid \mathcal{H}_1)}{p(\sigma \mid \mathcal{H}_0)}
    \overset{\mathcal{H}_1}{\underset{\mathcal{H}_0}{\gtrless}} \eta
\end{equation}

\noindent where $p(\sigma \mid \mathcal{H}_i)$ is the conditional likelihood
of the observed feature under hypothesis $\mathcal{H}_i$, and $\eta$ is the decision threshold chosen to satisfy a prescribed probability of false alarm $P_{FA} \leq \alpha$. This formulation maximizes the probability of detection $P_D$ subject to the false alarm constraint, making it well-suited for radar-based classification where missed detections and false alarms carry unequal costs \cite{jarabo2013radar}.

The discriminating feature is the spatial spread of the point cloud returns associated with each detected object, quantified by the 2D positional standard deviation of the cluster:

\begin{equation}
    \sigma = \sqrt{\frac{1}{N}\sum_{i=1}^{N}\left[(x_i - \bar{x})^2 +
    (y_i - \bar{y})^2\right]}
\end{equation}

\noindent where $(x_i, y_i)$ are the Cartesian coordinates of the $i$-th reflection point within the cluster, $(\bar{x}, \bar{y})$ is the cluster centroid, and $N$ is the total number of points in the cluster. Figure \ref{fig:base_classifier} shows the standard deviation distributions for passenger cars and motorcycles, where a decision threshold is originally set at their right-hand intersection (at 0.14~m). This threshold setting is not optimal since the associated error rate for object classification is $52.1\%$, which is substantially high. Hence, we propose to optimize the binary classifier via temporal averaging and a linear search for an optimal threshold.

\begin{figure}[t]
    \centering
    \includegraphics[width=0.75\columnwidth]{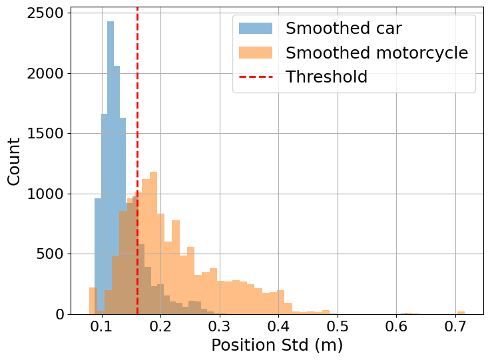}
    \caption{Class distributions of the 2D positional standard deviation 
    $\sigma$ after temporal averaging with window $W = 100$. The dashed line 
    indicates the optimal threshold $\eta^* = 0.16$ m.}
    \label{fig:new_classifier}
\end{figure}

\begin{figure}[h]
    \centering
    \includegraphics[width=0.8\columnwidth]{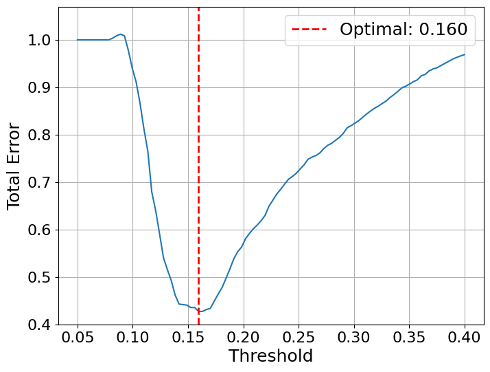}
    \caption{Total classification error rate as a function of the decision 
    threshold $\eta$. The optimal threshold $\eta^* = 0.16$ m minimizes the 
    error rate at $42.7\%$.}
    \label{fig:linear_search}
\end{figure}

\subsection{Improved Classifier}
As shown in Figure~\ref{fig:base_classifier}, the overlap between the two class distributions is substantial, which directly limits the separability of the base classifier. This overlap is a manifestation of the high temporal variance inherent in single-sweep radar measurements, where instantaneous point cloud returns are susceptible to multipath interference, target aspect-angle variation, and RCS fluctuations, all of which introduce 
stochastic noise into the per-frame $\sigma$ estimate.

To mitigate this, we apply a temporal averaging technique to suppress short-term measurement noise, while preserving the underlying kinematic signature of each tracked object. Specifically, a uniform moving average filter of window length $W = 100$ frames is applied to the $\sigma$ time-series of each track:
\begin{equation}
    \bar{\sigma}[n] = \frac{1}{W}\sum_{k=0}^{W-1} \sigma[n-k]
\label{equation:moving_average_filter}
\end{equation}
\noindent where $\bar{\sigma}[n]$ is the smoothed feature estimate at frame $n$. This operation is equivalent to a low-pass FIR filter with a rectangular impulse response, attenuating high-frequency variance components, while retaining the mean class signature. The smoothing reduces the variance of the passenger car distribution from $0.0022$ to $0.0013$, and that of the motorcycle distribution from $0.0100$ to $0.0067$, yielding tighter, more separable class-conditional distributions as illustrated in Figure~\ref{fig:new_classifier}.

With the smoothed feature distributions, we perform a linear search over the candidate threshold space to minimize the total classification error rate. The optimal threshold is found at $\eta^* = 0.16$~m, reducing the total error rate from $52.1\%$ to $42.7\%$, a reduction of $9.4$ percentage points. This improvement is significant in the context of traffic monitoring, where sustained misclassification of vulnerable road users, such as motorcyclists and cyclists, carries direct safety implications for downstream path planning and collision-avoidance systems. Figure~\ref{fig:linear_search} illustrates the error rate as a function of the candidate threshold, confirming the optimality of $\eta^*$.

%% file: 7_discussion.tex
\section{Discussion}
The proposed temporal averaging classifier achieves a total error rate of 42.7\%, a significant 9.4-percentage-point reduction from the 52.1\% baseline. These results show that temporal smoothing of the 2D positional standard deviation $\sigma$ is an effective and computationally lightweight strategy for improving radar-based vehicle classification. Nevertheless, the remaining error reflects the inherent challenge of single-feature discrimination when point cloud signatures overlap at extended ranges, and motivates the multi-modal fusion direction.
The binary scope of the classifier is a deliberate design choice grounded in the dataset's class distribution. Trucks and bicycles constitute only 0.24\% and 2.15\% of radar detections, respectively, making them statistically insufficient to derive reliable class-conditional models or evaluate multi-class performance meaningfully.

The moving average filter in Equation~\ref{equation:moving_average_filter} is computationally inexpensive and suitable for real-time deployment at the edge. However, the $W = 100$-frame window requires a track history before the smoothed feature estimate stabilizes, a cold-start latency that may be unacceptable for objects that enter the sensor field of view briefly or at high speed. Adaptive window lengths based on track age, or exponential moving averages with tunable decay constants, could mitigate this effect while preserving $\sigma$'s variance-reduction benefit.

An important consideration is the generalization of the proposed methodology beyond the dataset and sensing configuration considered in this work. The use of the spatial dispersion of radar point-cloud clusters and its temporal evolution is not inherently dependent on a specific road layout or traffic scenario, suggesting that the same processing methodology can be applied to other roadside deployments. However, the class-conditional distributions of $\sigma$, the decision threshold, and the most appropriate temporal window may depend on factors such as sensor configuration, installation geometry, target range, traffic composition, and environmental conditions. Consequently, the specific parameters obtained in this study should not be assumed to transfer directly to other deployments without further validation or recalibration. Evaluation across different locations and operating conditions is, therefore, required to determine the extent to which the observed class separability is preserved.

Several extensions follow naturally from this work. Incorporating additional discriminative features, such as radial velocity variance, RCS magnitude, and object heading, into a multi-feature classifier could improve separability beyond what $\sigma$ alone permits. In addition, camera-radar association and fusion could complement the radar-only classification and help resolve ambiguous cases.  
The use of commercial off-the-shelf sensors and computationally lightweight processing also makes the proposed platform straightforward to replicate, facilitating its evaluation in different roadside environments.

%% file: 8_conclusion.tex
\section{Conclusion}
This paper presented a roadside multi-sensor platform combining a 4D radar and a stereo depth camera for traffic monitoring under real urban conditions. The two sensing modalities provide complementary coverage: the radar captures extended trajectories over a substantially longer range, while the camera provides richer semantic information in the near-field region. Constructed from commercially available sensors and employing computationally efficient processing, this platform can be easily replicated for further roadside perception studies. Analysis of the collected data revealed an unexpectedly high proportion of motorcycle labels in the radar output, particularly for objects observed at longer ranges, even though passenger cars predominated in the traffic observed by the camera. 
To address this issue, a lightweight binary classification method based on the 2D spatial dispersion of the radar point-cloud clusters was implemented. By applying temporal averaging of this feature across individual object tracks, short-term variability was reduced, leading to a decrease in the reported classification error from 52.1\% to 42.7\%. Future work will explore adaptive temporal windows to reduce the cold-start latency of the smoothed feature, as well as the integration of additional radar features and camera data to further enhance classification reliability.